# REACT: Runtime-Enabled Active Collision-avoidance Technique for Autonomous Driving

Heye Huang[a], Hao Cheng[b], Zhiyuan Zhou[c], Zijin Wang[d], Qichao Liu[a], Xiaopeng Li[a] [*]

[a] *Department of Civil and Environmental Engineering, University of Wisconsin-Madison, WI 53706, USA*

[b] *School of Vehicle and Mobility, Tsinghua University, Beijing 100084, China*

[c] *Department of Financial Engineering, University of Southern California, CA 90089, USA*

[d] *Department of Civil, Environmental, and Construction Engineering, University of Central Florida, FL 32816, USA*

[*] *Corresponding author.*

*E-mail address: xli2485@wisc.edu.*

**Abstract**

Achieving rapid and effective active collision avoidance in dynamic interactive traffic remains a core challenge for autonomous driving. This paper proposes REACT (Runtime-Enabled Active Collision-avoidance Technique), a closed-loop framework that integrates risk assessment with active avoidance control. By leveraging energy transfer principles and human-vehicle-road interaction modeling, REACT dynamically quantifies runtime risk and constructs a continuous spatial risk field. The system incorporates physically grounded safety constraints such as directional risk and traffic rules to identify high-risk zones and generate feasible, interpretable avoidance behaviors. A hierarchical warning trigger strategy and lightweight system design enhance runtime efficiency while ensuring real-time responsiveness. Evaluations across four representative high-risk scenarios including car-following braking, cut-in, rear-approaching, and intersection conflict demonstrate REACT's capability to accurately identify critical risks and execute proactive avoidance. Its risk estimation aligns closely with human driver cognition ( i.e., warning lead time < 0.4 s), achieving 100% safe avoidance with zero false alarms or missed detections. Furthermore, it exhibits superior real-time performance (< 50 ms latency), strong foresight, and generalization. The lightweight architecture achieves state-of-the-art accuracy, highlighting its potential for real-time deployment in safety-critical autonomous systems.

**Keywords:** Autonomous driving, runtime risk assessment, collision avoidance, risk field modeling

## 1. Introduction

Achieving safe and efficient autonomous driving in dynamic and uncertain mixed-traffic environments remains a critical challenge for the widespread deployment of autonomous vehicles (Baccari et al., 2024; Lefèvre et al., 2014). Although continuous advances in perception and control technologies have significantly enhanced the functional safety of autonomous systems, ensuring operational safety under complex multi-agent interactions, where intentions are often difficult to predict, fundamentally hinges on the ability to anticipate potential risks and proactively avoid them (Guo et al., 2020; Wang et al., 2021). As a foundational component of proactive safety architectures, risk assessment aims to identify and quantify behaviors or spatial regions that may lead to future collisions, thereby guiding appropriate avoidance strategies. Typical applications, such as

Forward Collision Warning (FCW) and Autonomous Emergency Braking (AEB), have demonstrated substantial benefits in improving driving safety (Cicchino, 2023, 2017). However, existing systems often rely on rule-based heuristics or offline modeling for risk evaluation and avoidance control, lacking real-time, closed-loop validation in highly dynamic traffic. As a result, they fall short in capturing the heterogeneity and unpredictability of real-world threats (Grewal et al., 2024; Westhofen et al., 2023). When faced with high-risk scenarios, such systems frequently fail to answer fundamental questions such as *who* poses the risk, *when* and *where* it may occur, *why* it emerges, and *how* to avoid it effectively.

Existing research on risk quantification or threat assessment measures can be classified into four categories: (1) time and kinematics based measures, (2) probabilistic statistical measures, (3) behavior anomaly-based measures, and (4) field-based measures.

**Time and kinematics based indicators.** Among time-based indicators, Time to Collision (TTC), Time Headway (THW), Time to Brake (TTB), and Time to React (TTR) are commonly used (Ortiz et al., 2023; Reiffel, 2019). TTC is the most prevalent due to its simplicity and real-time performance, and is widely integrated into ADAS for longitudinal collision avoidance. Several extensions, such as Inverse TTC and Extended TTC, have been proposed to enhance stability and applicability. THW is more suitable for car-following scenarios and serves as a core reference for Adaptive Cruise Control (ACC). Although time-based indicators are efficient, they generally rely on constant-speed or constant-acceleration assumptions and lack the ability to model interactive multi-agent behavior. Kinematics-based metrics assess risk by evaluating deceleration and safe distance thresholds, including models from NHTSA, UC Berkeley, and acceleration-adjusted variants (Liu et al., 2025; Minguez and Montano, 2009). These approaches focus on whether a vehicle has sufficient space and capacity to avoid collisions under current conditions. However, they heavily depend on vehicle-specific physical properties and are limited in capturing the behavioral dynamics of other traffic participants. To address this, SafeCast incorporates Responsibility-Sensitive Safety (RSS) rules into multimodal motion prediction, using formal constraints to improve trajectory interpretability and safety (Liao et al., 2025). Yet, it remains difficult to generalize such methods to uncertain and diverse real-world behaviors and policy spaces.

**Probabilistic statistical measures.** To address these limitations, recent studies have introduced probabilistic statistical measures that estimate collision probability (CP) by modeling multi-trajectory distributions or uncertainty in future intentions (Schmerling and Pavone, 2017; Ward et al., 2014). These approaches often rely on Monte Carlo sampling to generate a variety of potential trajectories for traffic participants. By evaluating the probability of trajectory overlap, they provide risk estimates with greater behavioral diversity and dynamic adaptability. Wang et al. proposed a prediction-based risk metric by modeling vehicle state spaces, demonstrating that incorporating intention modeling improves the reliability and interpretability of risk predictions (Wang et al., 2024, 2022). Other research has developed risk assessment maps based on the spatial configuration of surrounding vehicles, defining the regions they are likely to occupy and translating them into risk zones linked to their positions and motions (Goli et al., 2018; Köhler, 2024). Given these expected occupancies, autonomous vehicles are required to respect and avoid such zones to reduce the risk of collision or conflict. However, these methods are often computationally intensive and difficult to deploy in real time within high-dimensional state spaces.

**Behavior anomaly-based measures**. These measures have gained increasing attention in recent years, aiming to expand the definition of risk by identifying behaviors that violate traffic rules or reflect cognitive anomalies in drivers (Li et al., 2021; Zhou and Zhong, 2020). Some studies have explored constructing context-specific risk metrics by mining operational guidelines embedded in traffic safety management rules (Rathore et al., 2023; Shi et al., 2022). Others have proposed identifying hazardous scenarios by comparing drivers' intended behaviors with the expected actions prescribed by road geometry and traffic regulations, particularly in intersection settings (Baccari et al., 2024; Huang et al., 2025). This line of research emphasizes uncovering latent threats from non-collision behaviors. However, these methods often rely on large-scale driving data analysis and still face technical challenges in behavior recognition accuracy and risk quantification for real-time deployment.

**Field-based measures.** These approaches have been widely explored. These methods construct risk potential fields guided by factors such as the position, velocity, and intent of surrounding vehicles to characterize hazardous regions in the traffic environment (Bounini et al., 2017; Li et al., 2020). Some studies explicitly represent the potential occupied space of other vehicles as ellipsoids or high-risk zones, using attractive or repulsive fields to guide avoidance strategies (Guerra et al., 2016; Kolekar et al., 2020). Wang et al. proposed a unified "driving safety field" model that incorporates human, vehicle, and road factors, enabling systematic quantification of driving risk in the presence of various obstacles, including dynamic agents (e.g., vehicles, pedestrians, cyclists) and static objects (e.g., curbs, speed bumps) (Wang et al., 2016). However, their approach does not explicitly consider road traffic risk from a traffic management perspective. Building on this, Huang et al. introduced driver intention modeling to enhance the dynamic adaptability of risk assessment (Huang et al., 2020). Different obstacles impose varying levels of threat and loss potential during collisions. Thus, integrating multiple types of traffic elements into a unified driving risk framework remains a challenging task. Moreover, most field-based methods rarely incorporate vehicle dynamics constraints during path planning, and are often affected by high parameter sensitivity, modeling uncertainty, and overfitting risks (Siradjuddin et al., 2024). Despite their strong ability to express spatial topology and support multidimensional risk assessment, these methods generally suffer from high computational complexity and poor real-time performance. Therefore, a lightweight and runtime-compatible multidimensional risk assessment model is still urgently needed for real-world deployment.

To address the above challenges, this paper proposes a lightweight, runtime-enabled active collision-avoidance technique (REACT). With human-vehicle-road interaction modeling, REACT considers how autonomous vehicles respond to future environmental changes by constructing behavior-aware risk fields. It dynamically identifies high-risk sources and proactively generates robust avoidance strategies, while ensuring physical feasibility, safety, and low-latency execution for real-world deployment. REACT enables accurate risk prediction and robust avoidance control in typical high-risk scenarios, such as car-following braking, cut-in, rear-approaching, and intersection conflict. Experimental results show that REACT achieves state-of-the-art avoidance performance with low latency and lightweight deployment, demonstrating the potential for real-time use in safety-critical autonomous driving systems. The contributions are as follows:

1) We propose a runtime-enabled multi-source risk field model that unifies behavior intention and physical modeling. By incorporating kinetic energy, velocity differences, and mass, it constructs a directionally adjustable and spatially decaying risk field to capture dynamic interactions.
2) We develop a grid-based risk map enabling directional risk attribution. Centered on the ego vehicle, the spatial discretization transforms the continuous risk field into a two-dimensional grid, integrating multi-source risk and road boundary constraints to identify dominant risk directions.
3) We introduce a lightweight, hierarchical warning and active avoidance strategy for real-time closed-loop control. Based on normalized runtime risk and adaptive thresholds, the system classifies risk into three levels and outputs interpretable warnings. Deployed on embedded platforms, it achieves high success rates across diverse safety-critical scenarios.

## 2. Framework

As illustrated in Fig. 1, the REACT targets dynamic interaction scenarios involving multiple traffic participants and establishes a runtime safety system with real-time risk assessment and active response capabilities. Based on a three-stage logic of risk modeling, grid-based risk map generation, and active avoidance, REACT forms an efficient perception, assessment, and control loop to manage high-risk situations in complex and dynamic traffic environments.

The framework consists of two main stages: 1) Runtime risk modeling. The system first formulates a kinetic-energy-based interaction risk field by incorporating the relative motion states of surrounding agents and embedding a spring potential field to reflect road boundary constraints. Within the ego vehicle's reachable area, a spatial grid discretization strategy is applied to build a localized risk map and compute directional risk statistics. From this map, a global risk value and directional risk indicators are extracted as inputs to downstream control modules. 2) Active collision avoidance strategy generation. Based on the risk map, REACT introduces dynamic risk thresholds and directional evaluation to support multi-level risk classification. It defines three levels of warning: Level 0 (Safe), Level 1 (Warning), and Level 2 (Emergency), corresponding respectively to safe driving, safe deceleration, and emergency braking or directional avoidance. By considering the runtime risk value, braking status, and velocity difference, the system triggers appropriate warnings (e.g., normal driving, deceleration recommended, emergency avoidance), and identifies the dominant risk direction. This enables the generation of adaptive, interpretable control outputs, such as throttle reduction, braking, or lateral evasive actions, to mitigate collisions safely.

REACT is modular and computationally lightweight, allowing deployment on embedded in-vehicle platforms for real-time active avoidance in various complex scenarios. By tightly coupling risk modeling with avoidance strategy generation, REACT overcomes the traditional separation of risk assessment and control, enabling intelligent, risk-driven decision-making at runtime.

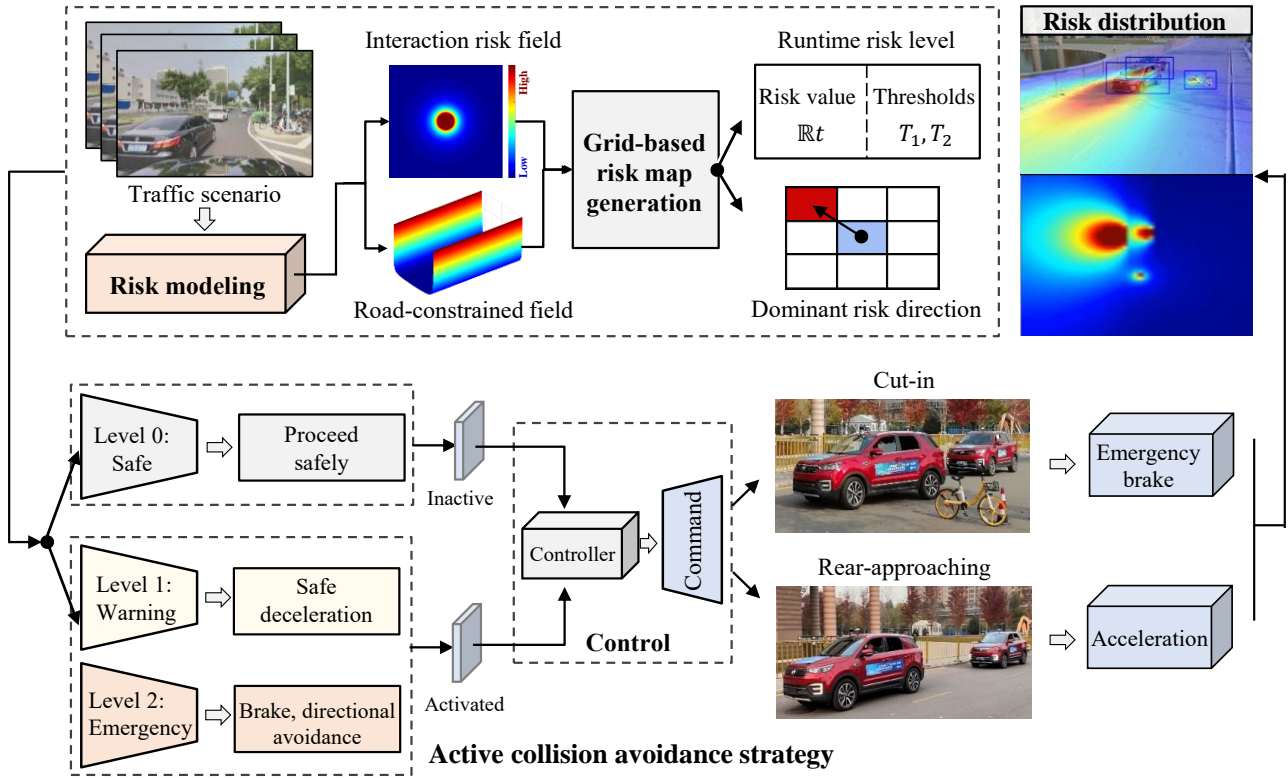

**Fig. 1.** Architecture of the REACT system for real-time risk mapping and hierarchical control

## 3. Method

*3.1 Multi-source risk modeling*

To support real-time risk assessment in the REACT system, this study establishes a foundational theorem for active safety, which provides the theoretical basis for quantifying driving risk dynamics and guiding timely avoidance decisions during runtime.

**Theorem 1 (Active Defensive Safety Strategy, ADS):** Consider an autonomous driving system initialized with a set of policies σ satisfying ADS. Given the existence of a true trajectory $\ell^*$ among all possible futures, if the system actively responds to $\ell^*$ and executes the corresponding strategy σ ($\ell^*$), it can progressively avoid risk and ensure collision-free driving.

Building upon this theoretical foundation, the REACT system introduces a runtime risk quantification and avoidance control mechanism as a practical embodiment of the proactive safety strategy. This mechanism enables continuous monitoring of the surrounding environment, dynamically evaluates potential hazards during operation, and triggers corresponding control strategies in a timely manner. It primarily comprises two steps: (1) construction of risk energy fields between the ego vehicle and multiple types of surrounding objects based on energy field theory; (2) the computation of a unified runtime risk value through spatial grid-based integration and normalization, which serves as the decision basis for activating context-aware control responses, including warnings, braking, or directional avoidance maneuvers.

### 3.1.1 Risk field modeling

In microscopic traffic systems, explicit or implicit dynamic interactions frequently occur between the ego vehicle and surrounding traffic participants, potentially leading to conflicts or collisions. Drawing an analogy to Coulomb forces or electrostatic potential fields between charged particles, the REACT framework models these interactions through a risk field based on the principle of "energy transmission." Each ego-conflict vehicle pair is abstracted as exerting a potential-like risk field acting upon the ego vehicle (Zheng et al., 2021). This field accounts for the dynamic interaction process across human-vehicle-road systems and incorporates key state features, including velocity, relative speed, and relative position.

For the ego vehicle, the intensity of each risk source is driven by three components: ego vehicle state: position $p_i = (x_i, y_i)$, velocity $v_i$, and mass $m_i$; participant vehicle state for $j \in J$: position $p_j = (x_j, y_j)$, velocity $v_j$, mass $m_j$ and type $\lambda_j$; relative state features: relative velocity vector $\boldsymbol{v}_{ij}$, its magnitude $\|\boldsymbol{v}_{ij}\|$, relative position $r_{ij}$, and the direction of the relative velocity $\theta_{ij}$.

$$\boldsymbol{v}_{ij} = \|\boldsymbol{v}_i - \boldsymbol{v}_j\| \tag{1}$$

$$r_{ij} = \sqrt{(x_i - x_j)^2 + (y_i - y_j)^2} \tag{2}$$

The angle $\theta_{ij}$ between the relative velocity direction and the relative position direction is defined as the angular deviation:

$$\theta_{ij} = \arccos(\frac{(x_i - x_j) \cdot (\boldsymbol{v_i} - \boldsymbol{v_j})_x + (y_i - y_j) \cdot (\boldsymbol{v_i} - \boldsymbol{v_j})_y}{\|\boldsymbol{r}_{ij}\| \cdot \|\boldsymbol{v}_{ij}\|}) \tag{3}$$

Unlike the Euclidean distance used in electrostatic potential fields, traffic interactions exhibit directional sensitivity due to vehicle motion. For the same spatial distance, an oncoming vehicle poses a greater risk than a vehicle approaching laterally. To capture this anisotropy, we adopt the Elliptical Distance, the velocity-direction-adjusted distance between the ego vehicle and a traffic participant is defined as:

$$\widetilde{r_{ij}}^2 = \frac{(x_i - x_j)^2}{a^2} + \frac{(y_i - y_j)^2}{b^2} \tag{3}$$

Typically, $a < b$, where the major axis is aligned with the velocity direction to reflect directional stretching. We set $a = r_{\text{long}}$, $a = k \cdot \| \boldsymbol{v_j} \|$, $k = 0.2s$, indicating that higher speeds lead to a broader longitudinal risk region. The lateral sensitivity is assumed constant, with $b = r_{\text{lat}} = 5$ m.

For each traffic participant $j \in J$, its objective risk potential at time $t$ is determined by its kinetic energy and can be expressed as:

$$U_j = \frac{1}{2} \lambda_j \cdot m_j \cdot \| \boldsymbol{v_j} \|^2 \tag{4}$$

**Interaction risk field $U_{ij}$.** Considering the interactive nature of traffic dynamics, each participant $j$ is abstracted as a risk source, centered around its kinetic energy, emitting a spatial risk energy density field. The composite risk energy field exerted on the ego vehicle $i$ by a traffic participant $j \in J$ at location $(x_i, y_i)$ is formulated as:

$$\begin{aligned} U_{ij}(x_i, y_i, t) &= U_j \cdot \left( 1 + \beta \cdot \cos(\theta_{ij}) \cdot \frac{\|\boldsymbol{v}_{ij}\|^2}{\|\boldsymbol{v}_j\|^2 + \epsilon} \right) \cdot \exp(-\widetilde{r_{ij}}^2) \\ &= \frac{1}{2} \lambda_j m_j \|\boldsymbol{v}_j\|^2 \cdot \left( 1 + \beta \cdot \cos(\theta_{ij}) \cdot \frac{\|\boldsymbol{v}_{ij}\|^2}{\|\boldsymbol{v}_j\|^2 + \epsilon} \right) \\ &\quad \cdot \exp\left( -\frac{(x_i - x_j)^2}{a^2} - \frac{(y_i - y_j)^2}{b^2} \right) \end{aligned} \tag{5}$$

where $\lambda_j$ denotes the participant type coefficient, reflecting the severity of potential collisions across different agent types (e.g., pedestrian, cyclist, passenger car, truck). Larger values indicate greater impact severity (e.g., truck = 1.5; pedestrian = 0.8). The term $m_j$ refers to the equivalent mass of participant $j$. Directional influence is modulated by $\cos(\theta_{ij})$: for oncoming vehicles, $\cos(\theta_{ij}) \approx 1$, the risk is amplified; for lateral targets, $\cos(\theta_{ij}) \approx 0$, the risk contribution is attenuated; and for receding targets, $\cos(\theta_{ij}) \approx -1$, the effect is controlled via a negatively correlated term. A small positive constant $\epsilon$, typically set to $10^{-6}$, is prevents division by zero. The exponential term $\exp(-\widetilde{r_{ij}}^2)$ models risk potential decay with respect to spatial proximity, analogous to a Gaussian kernel function: the closer participant $j$ is, the greater its influence.

In real-world deployment, when the ego vehicle approaches roadside boundaries or lane markings, an additional road-constrained field penalty term is introduced (Wang et al., 2016):

$$U_E(x_i, y_i, t) = \frac{1}{2} m_i \|\boldsymbol{v}_i\|^2 \cdot k_{\text{lane}} \cdot \lambda_{\text{lane}} \cdot \left( \frac{y_i - y_c}{y_{\text{max}}} \right)^2 \tag{6}$$

The dynamic road constraint term simulates the spring potential energy generated when the ego vehicle crosses lane boundaries, which activates when the ego vehicle deviates laterally beyond permissible bounds. A higher ego speed $\boldsymbol{v}_i$ results in stronger lateral deviation penalties. Here, $k_{\text{lane}}$ is a dimensionless road boundary stiffness coefficient (typically between 0.1 and 1), and $\lambda_{\text{lane}}$ is the lane line type coefficient (dashed line = 1; solid line = 1.5). The lateral deviation from the nearest boundary is denoted by $y_i - y_c$, where $y_c$ represents the longitudinal position of the lane edge. The coordinates of the left and right lane boundaries are $y_{c,r}, y_{c,l}$, respectively. The lane half-width $y_{\text{max}}$ is used for normalization (typically 1.75-2 m).

**Road-constrained field $U_{E,a}$.** By modeling lane boundaries as a symmetric dual-spring system, the total road-constrained field is expressed as:

$$U_{E,a}(x_i, y_i, t) = \frac{1}{2} m_i \|v_i\|^2 \cdot k_{\text{lane}} \cdot [\lambda_{\text{lane,r}} \cdot \left(\frac{y_i - y_{c,r}}{y_{\max}}\right)^2 + \lambda_{\text{lane,l}} \cdot \left(\frac{y_i - y_{c,l}}{y_{\max}}\right)^2] \tag{7}$$

**Total risk field $U_a$.** The total risk field acting on the ego vehicle is computed as the superposition of all individual participant fields and the road-constrained field:

$$U_a(x_i, y_i, t) = \sum_{j \in J} U_{ij}(x, y, t) + U_{E,a}(x_i, y_i, t) \tag{8}$$

**Field force $F_{ij}$.** The field force vector of the risk field $U_{ij}(x_i, y_i, t)$ is defined as the negative gradient of the field function, representing the direction of the steepest risk descent:

$$F_{ij}(x_i, y_i, t) = -\nabla U_{ij} = -\left(\frac{\partial U_{ij}}{\partial x_i} \vec{i} + \frac{\partial U_{ij}}{\partial y_i} \vec{j}\right) \tag{9}$$

The field force $F_{ij}$ points away from high-risk regions and drives the ego vehicle to avoid potential threats. Its magnitude $\|F_{ij}\|$ reflects the steepness of risk variation and serves as a basis for tuning control sensitivity. This force can be integrated into the trajectory planning phase to provide directional guidance, analogous to the motion of electrons in an electrostatic potential field.

The partial derivative in the $x$-direction is expanded as:

$$\frac{\partial U_{ij}}{\partial x_i} = -\lambda_j m_j \|v_j\|^2 \cdot \left(1 + \beta \cdot \cos(\theta_{ij}) \cdot \frac{\|v_{ij}\|^2}{\|v_j\|^2 + \epsilon}\right) \cdot \frac{(x_i - x_j)}{a^2} \cdot \exp(-\widetilde{r}_{ij}^2) \tag{10}$$

The partial derivative in the $y$-direction is expanded as:

$$\frac{\partial U_{ij}}{\partial y_i} = -\lambda_j m_j \|v_j\|^2 \cdot \left(1 + \beta \cdot \cos(\theta_{ij}) \cdot \frac{\|v_{ij}\|^2}{\|v_j\|^2 + \epsilon}\right) \cdot \frac{(y_i - y_j)}{b^2} \cdot \exp(-\widetilde{r}_{ij}^2) \tag{11}$$

This force represents the attractive or repulsive influence exerted by participant $j$ at a spatial point $(x_i, y_i)$, and its direction aligns with the gradient descent of the risk field. When multiple traffic participants are present, the total field force is obtained by vector superposition:

$$\begin{aligned} F_a(x_i, y_i, t) &= -\nabla U_a = \sum_{j \in J} F_{ij}(x_i, y_i, t) - \nabla U_{E,a}(x_i, y_i, t) \\ &= \sum_{j \in J} \left(-\frac{\partial U_{ij}}{\partial x_i} \vec{i} - \frac{\partial U_{ij}}{\partial y_i} \vec{j}\right) + (0, -\frac{\partial U_{E,a}}{\partial y_i} \vec{j}) \end{aligned} \tag{12}$$

3.1.2 Spatial discretization and grid-based risk map generation

To enable real-time modeling and visualization of local risk distribution, this paper proposes a spatial discretization strategy centered on the ego vehicle. Based on the risk energy field, a structured, two-dimensional grid-based risk map is constructed, from which a unified runtime risk value is extracted to guide avoidance strategies.

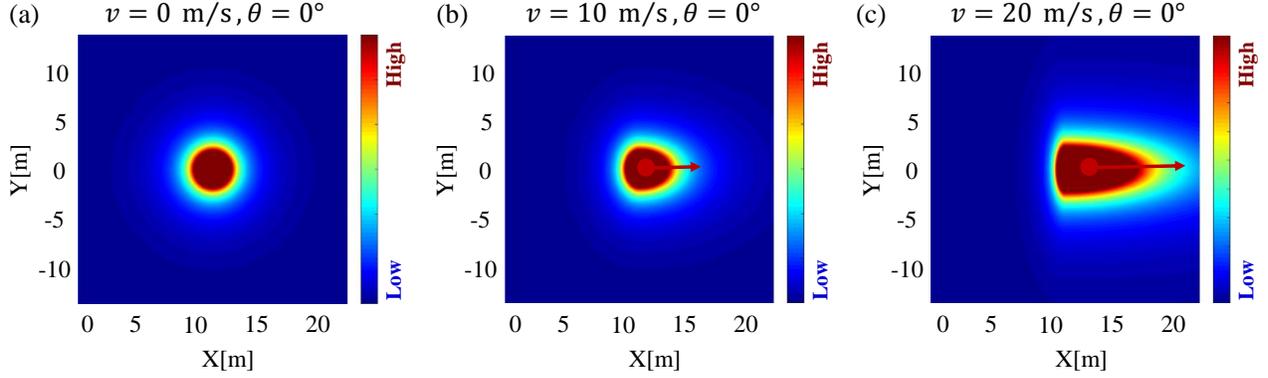

Fig. 2. Risk field visualization. (a) Field without velocity; (b) Field with low velocity; (c) Field with high velocity.

1) Local reachable space definition: At each system time step $t$, a local reachable space $\Omega$ is defined around the ego vehicle, covering critical front, rear, and lateral areas. To ensure efficiency, $\Omega$ is discretized into eight representative positions (front, rear, front-left, mid-left, rear-left, front-right, mid-right, rear-right), and further divided into an $m \times n$ grid $\{c_{i,j}\}$, with each cell centered at $(x_{i,j}, y_{i,j})$. The discretization is applied only within the Region of Interest (ROI) to avoid excessive grid refinement over the global scene, ensuring both spatial coverage and real-time performance.

2) Risk intensity calculation at grid centers: For each grid cell $c_{i,j}$, the cumulative field force exerted by all surrounding traffic participants $j$ is calculated based on their dynamic positions at time $t$. This interaction-based potential is denoted as:

$$\mathcal{R}_{i,j}^t = U_a(c_{i,j}, t) = U_a(x_{i,j}, y_{i,j}, t) \tag{7}$$

The computed values across the entire spatial grid collectively form a two-dimensional risk intensity matrix:

$$\mathcal{R}_t = \begin{bmatrix} \mathcal{R}_{1,1} & \cdots & \mathcal{R}_{1,n} \\ \vdots & \ddots & \vdots \\ \mathcal{R}_{m,1} & \cdots & \mathcal{R}_{m,n} \end{bmatrix} \tag{8}$$

3) Grid integration and normalization: To simplify global decision-making and facilitate downstream control operations, the system performs normalized aggregation over the entire risk map, resulting in a unified scalar runtime risk value $\mathbb{R}_t$:

$$\mathbb{R}_t = \frac{1}{m \cdot n} \sum_{i=1}^{m} \sum_{j=1}^{n} \mathcal{R}_{i,j}^t \tag{9}$$

Simultaneously, the map is partitioned into multiple semantically meaningful directional subregions (e.g., front-left, rear-right), each associated with a specific angular sector. For each subregion $d$, the average risk energy is calculated as:

$$\overline{\mathbb{R}_d} = \frac{1}{|C_d|} \sum_{(i,j) \in C_d} \mathcal{R}_{i,j}^t \tag{10}$$

where $C_d$ denotes the set of grid cells corresponding to direction $d$, (e.g., front-left corresponds to a sector with $\theta \in [45°, 90°]$). The final outputs for the warning system include:

- Current risk level: determined by comparing the runtime risk value $\mathbb{R}_t$ against thresholds $T_1, T_2$;
- Dominant risk direction: $d^* = \arg\max_d (\overline{\mathbb{R}_d})$, providing semantic interpretability;
- Recommended strategy type: e.g., "High risk detected in the front-left, deceleration advised," or "Approaching vehicle from the rear-right, maintain lane."

Lastly, a grid-based risk map is generated, in which each grid cell $\{c_{i,j}\}$ is assigned an average directional risk energy $\overline{\mathbb{R}_d}$ for real-time risk assessment.

Therefore, we model the driving risk as a field-based distribution. For a single agent, without considering speed and direction, the resulting risk forms a circular field, as shown in Fig. 2(a). When speed is included, risk concentrates along the motion direction and increases with speed (Fig. 2(b), 2(c)).

### 3.1.3 Model comparison

**Table 1** Comparison of risk assessment models

| Model | | Limitation | Summary |
|---|---|---|---|
| **TTC (Time to Collision)** 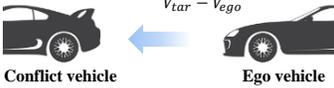 | $TTC = \dfrac{d}{v_l - v_e}$<br>$d$: inter-vehicle distance; $v_l$, $v_e$: the speeds of leading and ego vehicle. | Only considers longitudinal risks, ignores lateral interactions. | Single-dimensional (1D) risk assessment, insufficient for complex real-world scenarios |
| **THW (Time Headway)** 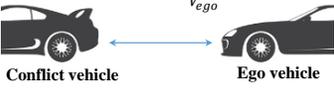 | $TTC = \dfrac{d}{v_e}$<br>$d$: inter-vehicle distance; $v_e$: the speed of ego vehicle. | | |
| **RSS (Responsibility-Sensitive Safety)** 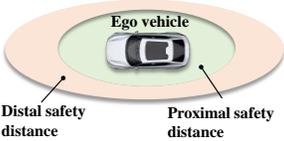 | $RSS = a + \dfrac{v^2}{2b}$<br>$a$: acceleration; $v$: velocity, and $b$: braking intensity. | Scenario-specific tuning, sensitive to environmental changes. | 2D safety model, limited generalization in multi-agent settings |
| **DRF (Driver risk Field)** 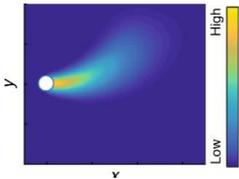 | $F = -\nabla V$<br>Based on the force field model, $V$: the potential function. | Ineffective in complex human-vehicle-road interactions. | Limited for complex multi-agent, real-world scenarios. |
| **REACT** 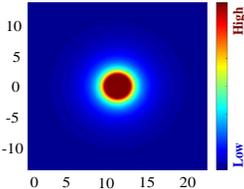 | $F_a(x_i, y_i, t) = -\nabla U_a$<br>Risk assessment based on a two-dimensional distribution field. | - | Captures multi-factor risk; generalizable to dynamic environments |

To clarify the strengths and limitations of existing risk assessment methods in autonomous driving, Table 1 provides a comparative overview of five representative models: Time-to-Collision (TTC), Time-Headway (THW), Responsibility-Sensitive Safety (RSS) (Hasuo, 2022), Driver Risk Field (DRF) (Kolekar et al., 2020), and the proposed REACT framework. These models differ in their risk representation, dimensionality, and adaptability to complex traffic scenarios. Traditional metrics such as TTC and THW focus solely on longitudinal dynamics, offering limited insight into lateral interactions. RSS introduces a two-dimensional safety buffer but relies on scenario-specific parameters, which may lead to misjudgment under dynamic conditions. DRF introduces a force-field abstraction but lacks robustness in multi-agent contexts. In contrast, REACT adopts a field-based risk modeling approach, enabling the integration of multi-source factors (e.g., vehicle dynamics, road geometry, and surrounding agents) into a unified probabilistic framework. This formulation improves risk generalization and supports real-time decision-making in diverse traffic scenarios.

*3.2 Active collision avoidance strategy generation*

Following the construction of the risk map and computation of the runtime risk value, the REACT framework evaluates the current driving state and generates interpretable, direction-aware safety strategies. As a key interface to the driver or higher-level decision modules, it outputs a warning level and recommended action, rather than direct control commands, ensuring controllability and effective collaboration.

3.2.1 Warning trigger strategy

The REACT system determines the current risk level and issues graded warnings based on the runtime risk value $\mathbb{R}_t$, the braking state $S_{\text{brake}} \in [0,1]$, and the relative velocity $\Delta\Delta v = v_{\text{der}} - v_i$.

1) Dynamic threshold adjustment. The base risk thresholds are set as $T_1 = 0.3$ and $T_2 = 0.7$, and are dynamically adjusted according to the driving state:

$$\begin{cases} T_1' = T_1\left(1 + \frac{\Delta v}{30}\right) \\ T_2' = T_2(1 - S_{\text{brake}}) \end{cases} \quad (11)$$

When the target speed is significantly higher than the current speed ($\Delta v > 0$), early warnings are necessary, and the lower threshold $T_1'$ is increased. During braking ($S_{\text{brake}} > 0$), the upper threshold $T_2'$ is reduced to enable earlier detection of high-risk scenarios.

2) Graded Warning Logic. Based on the adjusted thresholds $T_1'$ and $T_2'$, the runtime risk value $\mathbb{R}_t$, and the directional subregion risk $\overline{\mathbb{R}_d}$, a three-level warning is triggered as shown in Table 2:

**Table 2** Runtime risk level classification and warning strategy mapping

| Condition | Risk Level | Warning Output |
|---|---|---|
| $\mathbb{R}_t < T_1'$ or $S_{\text{brake}} = 1$ | Level 0: Safe | Strategy: Proceed safely |
| $T_1' \leq \mathbb{R}_t < T_2'$, $S_{\text{brake}} = 0$ | Level 1: Warning | Strategy: Deceleration |
| $\mathbb{R}_t \geq T_2'$, $S_{\text{brake}} = 0$ | Level 2: Emergency | Strategy: Emergency braking + directional avoidance (Type $j$ approaching rapidly from direction $d$) |

By integrating environmental perception, the active avoidance module in REACT can, upon triggering a warning, clearly indicate the type of risk source (e.g., pedestrian, vehicle, cyclist), its direction (e.g., front-left, directly ahead, rear), and the recommended response (e.g., decelerate, brake, change lane). This enables the delivery of transparent, interpretable risk alerts and strategy outputs. The mechanism is further linked to the human-machine interface, allowing real-time semantic warnings when high-risk targets are detected, such as crossing pedestrians or laterally merging vehicles. Representative outputs include: "Pedestrian crossing rapidly from the front-left, immediate deceleration recommended," or "Vehicle approaching from the rear-right, maintain lane and drive cautiously."

3.2.2 Runtime risk assessment and active warning level detection

To support real-time risk assessment and early warning decisions, we implement the REACT algorithm as shown in Algorithm 1. The algorithm first constructs a grid-based risk map centered on the ego vehicle by aggregating the contributions from all surrounding traffic participants and dynamic road constraints. Then, it calculates the global and directional risk levels to identify the dominant threat sector. Finally, it outputs a three-level advisory command (normal, cautionary, emergency) based on dynamic threshold adjustment. This modular design ensures that REACT can provide interpretable and proactive safety recommendations with low computational overhead, making it suitable for onboard deployment.

---

**Algorithm 1** REACT: runtime risk assessment and active warning level detection

---

**Input:** Ego vehicle state $p_i = (x_i, y_i), v_i, m_i$, traffic participants set: $j \in J$, each with state $p_j, v_j, m_j, \lambda_j$, parameters: $\beta$ (direction gain), $\epsilon$ (small constant), $k_{\text{lane}}$ (lane stiffness), risk thresholds: $T_1, T_2$

**Output:** Risk matrix $\mathcal{R}_{i,j}^t$ for each grid cell, global risk level: $\mathbb{R}_t$ = average($\mathcal{R}_{i,j}^t$), directional risks: $\overline{\mathbb{R}_d}$, warning level: L ∈ {0: low, 1: medium, 2: high}, Command: A

1: Initialize risk matrix R ← zeros(m, n)

2: for each grid cell $c_{i,j} \in \Omega$ do

3:   for each $j \in J$ do

4:     Compute relative velocity $\mathbf{v}_{ij}$, angle $\theta_{ij}$

5:     Compute elliptical distance $\widetilde{r}_{ij}^{\,2} \leftarrow (x_i - x_j)^2/a^2 + (y_i - y_j)^2/b^2$

6:     Compute $U_{ij} \leftarrow 0.5\, \lambda_j m_j \|\mathbf{v}_j\|^2 \cdot \left(1 + \beta \cdot \cos(\theta_{ij}) \cdot \|\mathbf{v}_{ij}\|^2 / (\|\mathbf{v}_j\|^2 + \epsilon)\right) \cdot \exp(-\widetilde{r}_{ij}^{\,2})$

7:     $\mathcal{R}_{i,j}^t$ += $U_{ij}$

8:   end for

9:   Add road potential $U_{E,a} \leftarrow 0.5\, m_i \|\mathbf{v}_i\|^2 \cdot k_{\text{lane}} \cdot \left[\lambda_{\text{lane,r}} \cdot \left(\frac{y_i - y_{c,r}}{y_{\max}}\right)^2 + \lambda_{\text{lane,l}} \cdot \left(\frac{y_i - y_{c,l}}{y_{\max}}\right)^2\right]$

10:   $\mathcal{R}_{i,j}^t$ += $U_{E,a}$

11: end for

12: Compute global risk $\mathbb{R}_t \leftarrow (1/mn) \cdot \sum \mathcal{R}_{i,j}^t$

13: For each direction $d \in$ {FR, R, RR, B, RL, L, FL, F}

14:   Compute sector risk $\overline{\mathbb{R}_d}$

15:   Dominant direction $d^* \leftarrow \arg\max_d (\overline{\mathbb{R}_d})$

16: Update thresholds: $T_1' \leftarrow T_1(1 + \Delta v/30)$, $T_2' \leftarrow T_2(1 - S_{\text{brake}})$

17: Determine warning level L and command action A

18: if $\mathbb{R}_t < T_1'$ or $S_{\text{brake}} = 1$, then L ← 0, A ← "Normal driving"

19: elif $T_1' \leq \mathbb{R}_t < T_2'$, $S_{\text{brake}} = 0$, then L ← 1, A ← "Reduce speed to avoid risk in " + $d^*$

20: else L ← 2, A ← "Emergency action toward opposite of " + $d^*$

21: **Return:** risk map $\mathcal{R}_{i,j}^t$, global risk $\mathbb{R}_t$, directional risks $\overline{\mathbb{R}_d}$, direction $d^*$, level L, command A

## 4. Experiments

To evaluate REACT in high-risk scenarios, we first validated it on real-world datasets, then implemented a lightweight runtime control system and deployed it on a real vehicle. The modular architecture enables real-time perception input to the REACT behavior module for end-to-end risk assessment and active avoidance.

*4.1 Comparative validation on real-world datasets*

To validate the effectiveness of the proposed REACT framework, we conduct a comparative study against five representative risk assessment models listed in Table 1: Time-to-Collision (TTC), Time Headway (THW), Responsibility-Sensitive Safety (RSS), Drive Risk Field (DRF), and REACT. Experiments are performed on the highD dataset, a real-world highway driving dataset collected in Germany, which captures complex vehicle interactions such as cut-ins, lane changes, and lateral merges. The highD dataset contains over 110,000 trajectories across 60 highway scenarios, offering high-precision data on position, speed, acceleration, and lane-level behavior. Its multi-agent dynamics and diverse conflict patterns make it ideal for evaluating risk assessment models under realistic conditions.

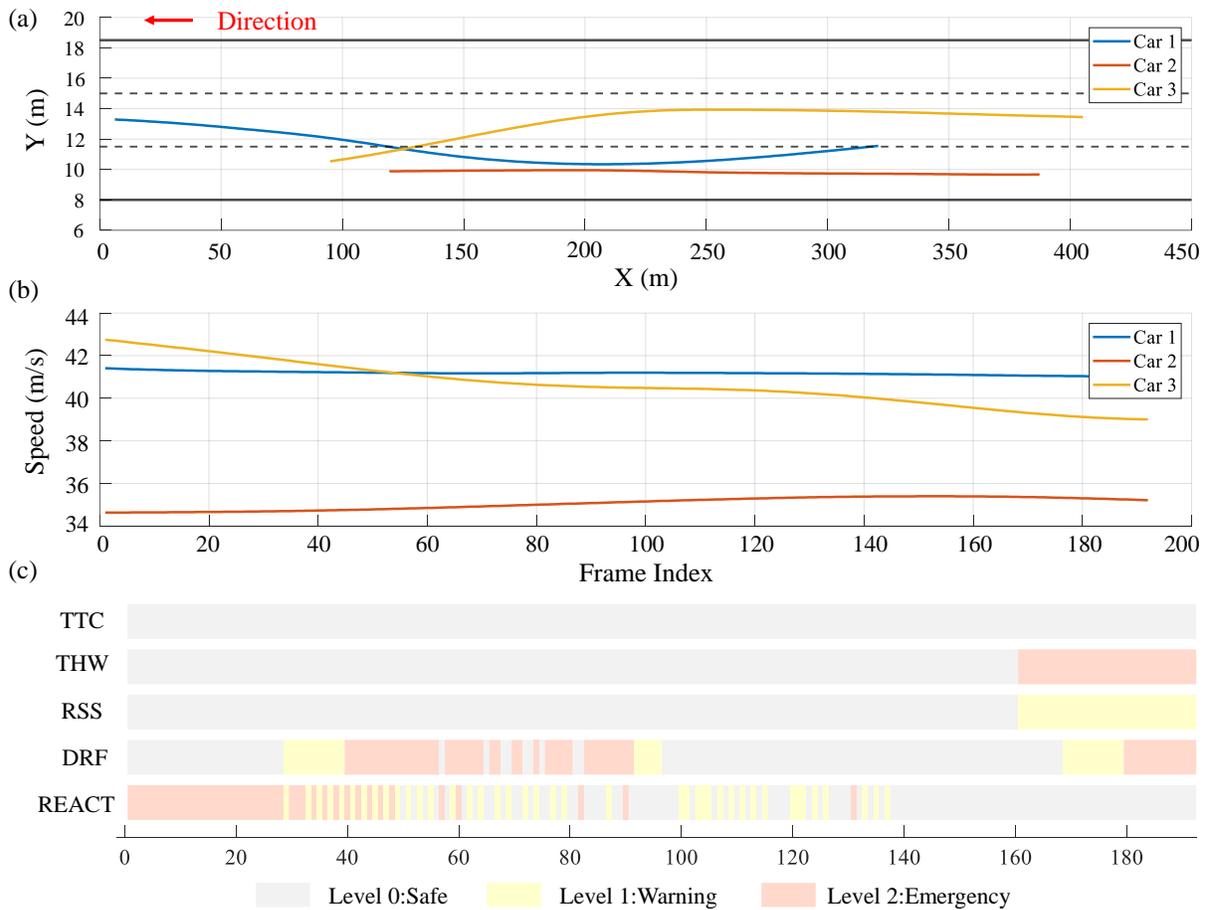

**Fig. 3.** Comparative risk assessment in a highway lane-change scenario using the highD dataset

Fig. 3 illustrates the risk assessment results of five models, TTC, THW, RSS, DRF, and REACT, in a highway cut-in scenario using the highD dataset. (a) shows the spatial trajectories of three vehicles: Car3 (yellow) is executing a cut-in from the right, Car2 (red) is the ego vehicle, and Car1 (blue) changes to the left lane during the scenario; (b) presents the speed profiles, where Car3 merges into the ego lane with a relatively lower speed; (c) compares model risk outputs. TTC, THW, and RSS are not sensitive to lateral cut-in risk and fail to detect any critical situation during the merging process, only triggering high-risk warnings after the cut-in is complete due to reduced longitudinal spacing. Both DRF and REACT are able to detect the lateral risk during the cut-in and activate warnings accordingly. However, DRF shows delayed response due to its lack of modeling for rear threats before the cut-in. While REACT accurately captures interaction dynamics between vehicles. For instance, after Car3 completes the cut-in, it maintains a higher speed than the ego vehicle, and REACT does not unnecessarily re-trigger the alarm, demonstrating superior directional sensitivity and real-time robustness in recognizing compound, multi-source interaction risks.

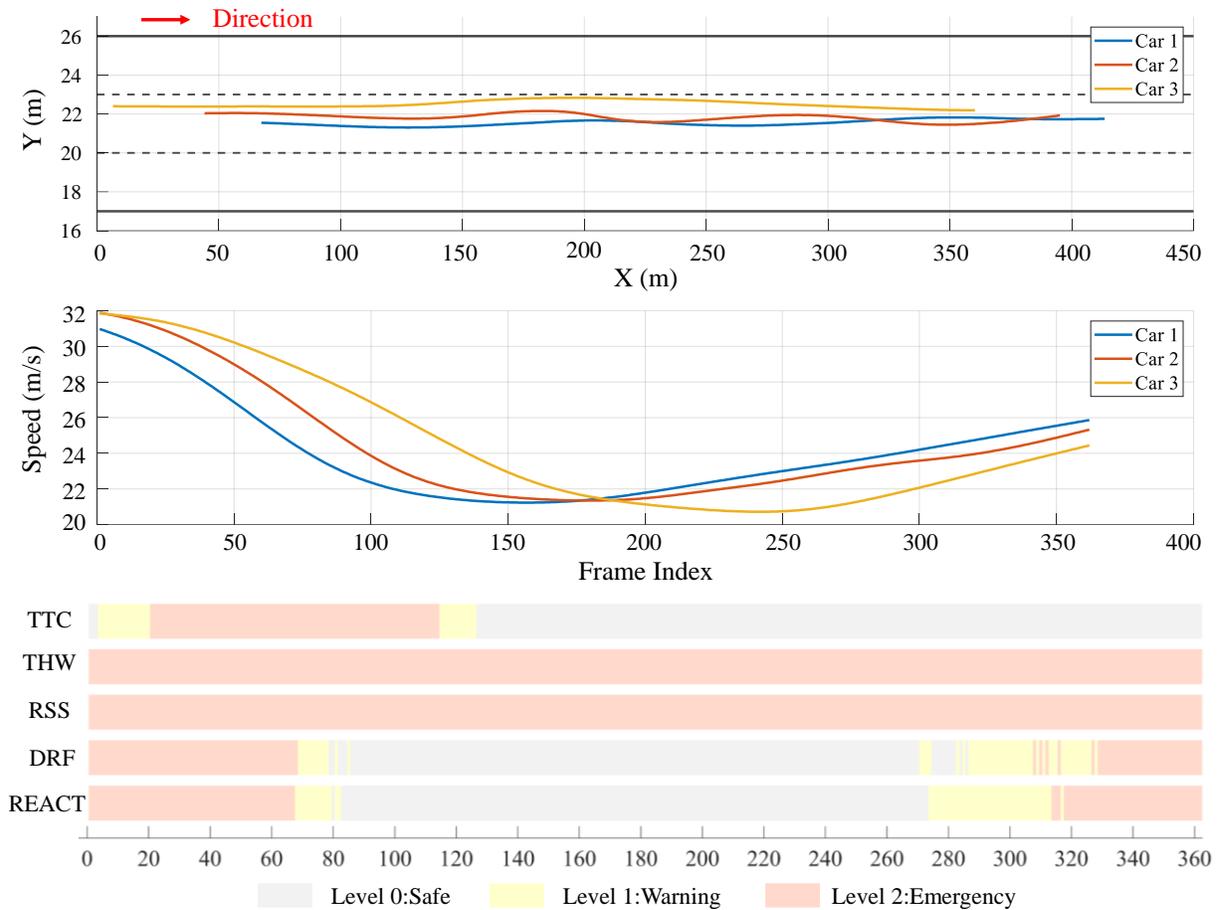

**Fig. 4.** Comparative risk assessment in a highway car-following scenario using the highD dataset

Fig. 4 compares risk assessment performance during a car-following deceleration scenario. (a) shows spatial trajectories, where the ego vehicle Car2 (red) initially follows Car1 (blue) with Car3 (yellow) behind; (b) illustrates speed profiles, with Car1 suddenly decelerating and triggering a chained response in Car2 and Car3; (c) presents the model-identified risk levels. The TTC model generates a brief emergency alert at the beginning due to close headway, but issues no further warnings during the following phase as the lead vehicle maintains a higher speed than the ego vehicle. In contrast, both THW and RSS respond strongly at the onset of deceleration but persistently output high-risk levels even after the interaction stabilizes, lacking the ability to recognize conflict resolution. DRF and REACT outperform others by accurately capturing the evolution of longitudinal risk based on dynamic speed and distance changes. They provide timely warnings during the critical deceleration phase and deactivate alerts once the risk subsides, demonstrating robust adaptability and effective graded risk perception.

*4.2 On-vehicle risk assessment experiments*

4.2.1 Hardware platform

The experimental platform consists of two Changan CS55 SUVs and one COMS electric vehicle. As shown in Fig. 5, all vehicles are drive-by-wire enabled and equipped with LiDAR, onboard IMU, differential GPS, dashcams, and industrial-grade computing units, supporting high-frequency data acquisition and closed-loop real-time control. The platform allows coordinated multi-vehicle operation, providing a comprehensive hardware-software foundation for validating runtime risk assessment and active avoidance strategies.

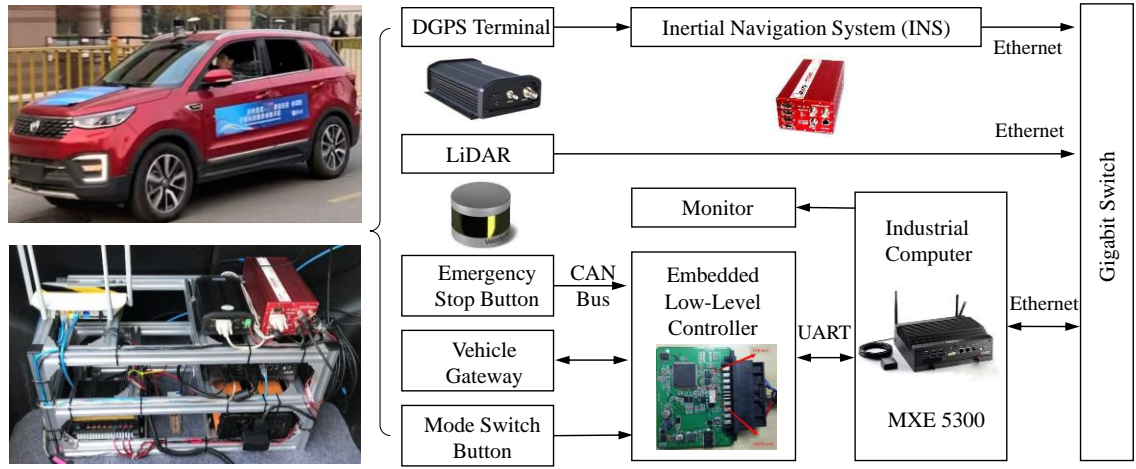

**Fig. 5.** Autonomous vehicle platform and physical architecture

4.2.2 Experimental design

The experimental design includes four representative high-risk scenarios—car-following braking, cut-in, rear-approaching, and intersection conflict (Fig. 6), to evaluate REACT's real-time risk assessment and control strategy under complex interactions. Vehicle states (e.g., position, speed, acceleration, braking) are recorded, and REACT-generated risk values trigger directional prompts based on the relative positions of conflict vehicles. When the risk exceeds a predefined threshold, the system issues voice alerts (e.g., "Danger ahead on the right, please brake") to support driver avoidance. Scenario details are summarized in Table 3.

**Table 3** Descriptions of typical high-risk driving scenarios

| Scenario | What (State) | Where | How |
|---|---|---|---|
| Car-following braking | Frontal risk due to sudden braking. Ego: ~19 km/h; lead vehicle: 18 km/h; gap: ~15 m; brakes after 5-8 s. | Urban straight road Fig. 5(a) | Keep distance, brake early |
| Cut-in | Lateral and frontal risk from adjacent lane cut-in. Ego: 20 km/h; cut-in vehicle: 20 → 25 km/h from 10 m ahead due to front obstacle. | Merging segment Fig. 5(b) | Monitor risk area, slow down |
| Rear-approaching | Rear-end risk from fast-approaching vehicle. Ego: 15 km/h; rear vehicle: 15 → 20 km/h from 20 m behind after 5 s. | Low-speed road Fig. 5(c) | Maintain gap, detect rear risk |
| Intersection conflict | Lateral collision risk at unsignalized crossing. Both vehicles: 20 km/h; 40 m from intersection, approaching from perpendicular directions. | Unsignalized intersection Fig. 5(d) | Slow down, monitor risk area |

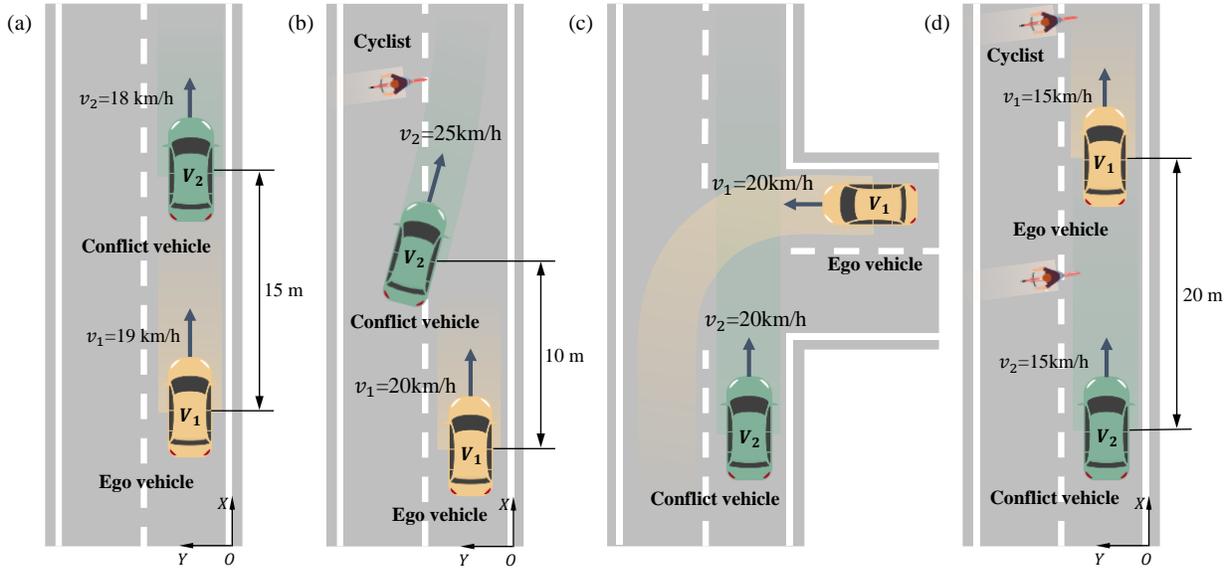

**Fig. 6.** Representative high-risk driving scenarios: (a) Car-following braking, (b) Cut-in, (c) Rear-approaching, (d) Intersection conflict

*4.3 Experimental results and analysis*

Real-world experiments are evaluated using false alarm rate, miss rate, warning time, prediction range, and braking timing. Braking timing, indicating risk assessment outcomes, is compared with ground truth from drivers to assess risk identification accuracy.

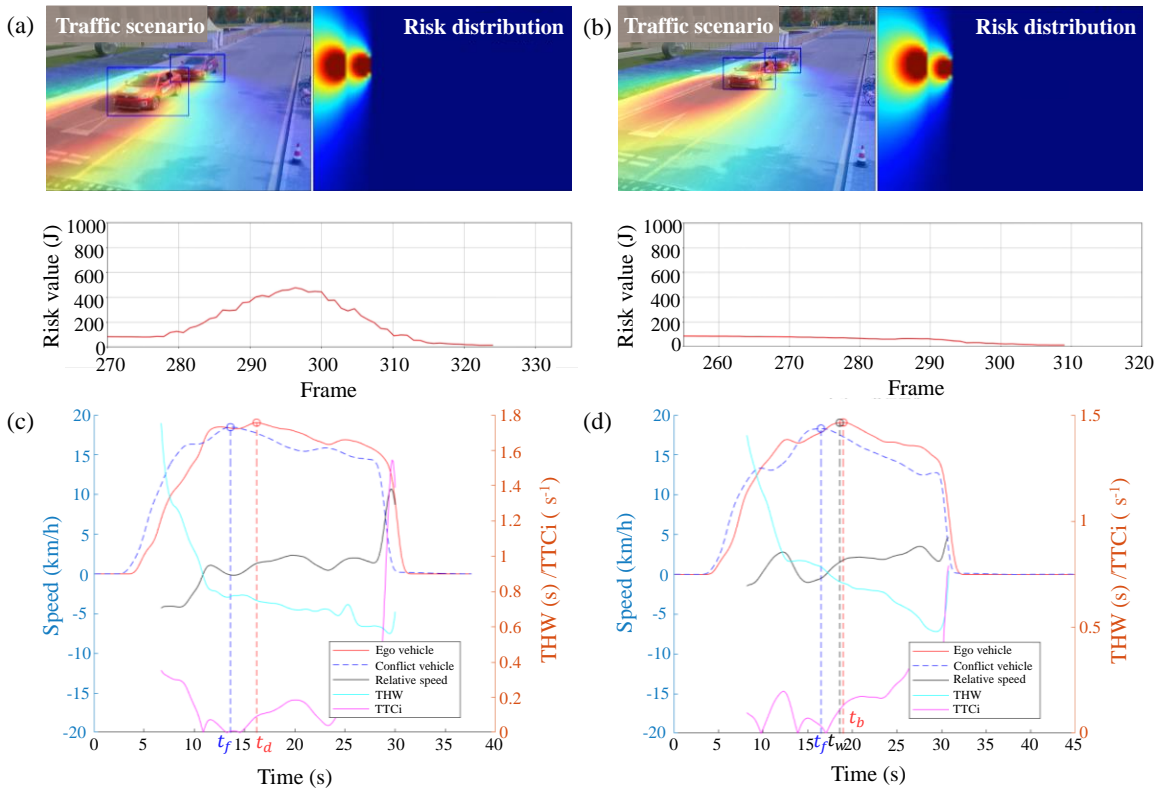

**Fig. 7.** Car-following braking scenario: comparison with and without warning. (a) (c) Risk distribution and driver behavior without warning; (b) (d) Risk distribution and driver behavior with REACT-based warning.

As shown in Fig. 7-10, four high-risk scenarios were tested under two conditions: without warning and with REACT-based warning. Each figure compares the risk distribution and driver behavior between the two settings. In each figure, subplots (a) and (c) show results without warning, while (b) and (d) show results with warning assistance. In subplots (a) and (b), the top-left image shows the actual traffic scene; the top-right shows the computed risk heatmap; and the bottom displays the dynamic risk curve, with energy values (in joules) indicating risk magnitude. In subplots (c) and (d), driver behavior is visualized: the red solid line indicates the ego vehicle's speed; the blue dashed line shows the conflict vehicle's speed. Key time points include: $t_f$: moment when the conflict vehicle changes state (e.g., sudden deceleration or acceleration; blue circle); $t_d$: reaction time of the driver without warning (red circle); $t_w$: time when REACT triggers a warning (black circle); $t_b$: driver reaction time with warning (red circle). A greater relative speed difference between vehicles indicates increased interaction intensity. Metrics such as THW and TTC$^{-1}$ serve as references for quantifying the objective conflict severity of each scenario.

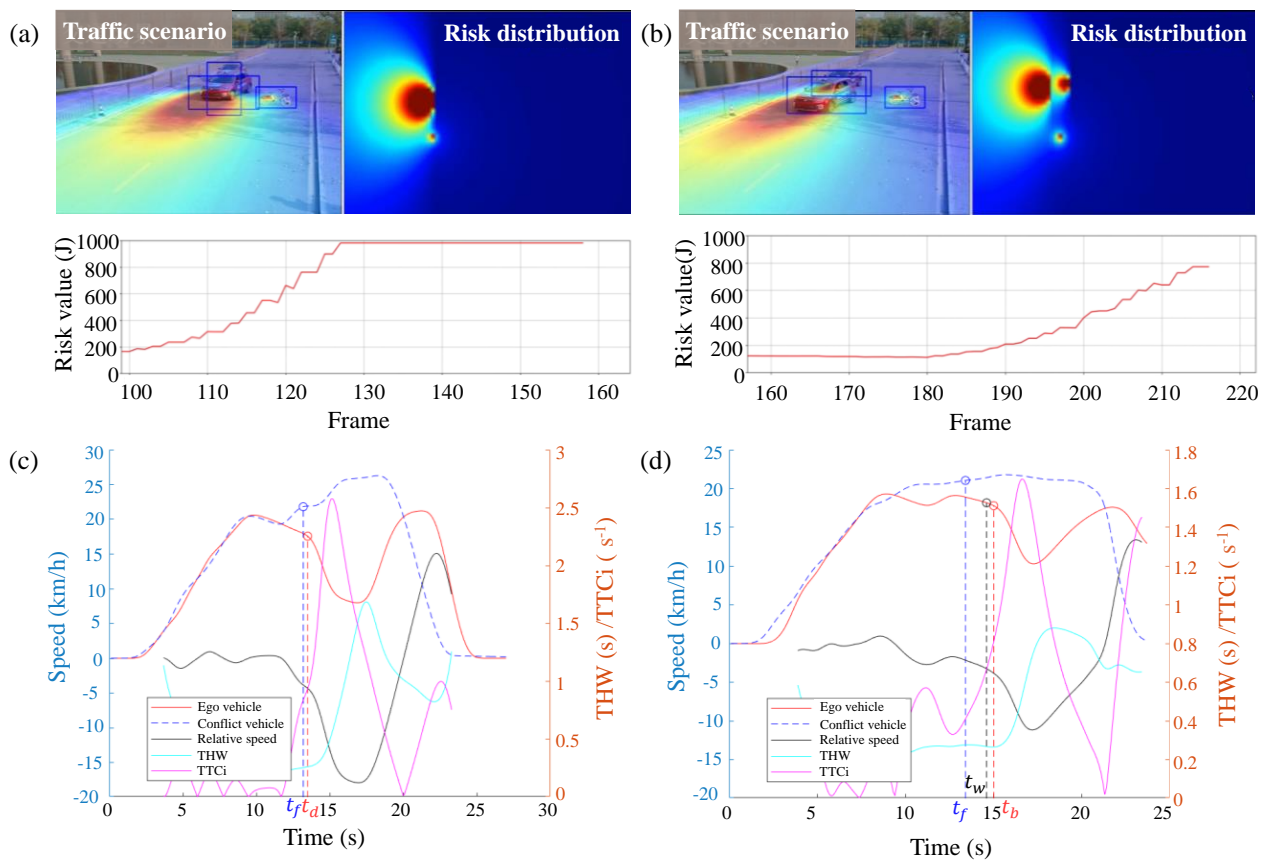

**Fig. 8.** Cut-in scenario: comparison with and without warning. (a) (c) Risk distribution and driver behavior without warning; (b) (d) Risk distribution and driver behavior with REACT-based warning.

In the car-following braking scenario, Fig. 7 illustrates the contrast between trials without warning (a, c) and those with REACT-enabled warning assistance (b, d). Strong consistency is observed between the driver's spontaneous responses and REACT's issued alerts, indicating the system's ability to reflect human-like risk assessment. Both active and warning-assisted responses reduce driving risk. The THW and TTC$^{-1}$ exhibit a typical trend, first increasing to indicate rising urgency, then decreasing as the conflict is gradually resolved. Notably, when REACT warnings are active, the ego vehicle initiates deceleration earlier and maintains a larger following distance, thereby minimizing the overall risk level and improving safety margins. In the cut-in scenario (Fig. 8), the ego vehicle decelerates to avoid a lateral intrusion. REACT detects the risk of lateral encroachment in advance by analyzing the relative speed and distance between vehicles, issuing directional risk alerts that guide timely braking and promote safer merging interactions.

In the rear-approaching scenario (Fig. 9), REACT identifies rearward threats earlier than the driver and recommends forward acceleration to increase inter-vehicle spacing. Although this maneuver briefly elevates risk due to speed-up, the subsequent growth in separation distance leads to a rapid risk reduction. Without such alerts, drivers exhibit delayed reactions, resulting in prolonged exposure to elevated risk. Similarly, in the intersection conflict scenario (Fig. 10), the algorithm detects both lateral and longitudinal threats that may be obscured by environmental blind spots. This early perception capability enables the system to generate timely warnings, assisting drivers in avoiding imminent collisions and effectively reducing intersection-related driving risks.

As shown in (c) and (d), driver response timing with and without warnings is compared across scenarios. Notably, in warning-assisted cases, the driver typically waits for the system alert before acting, which may result in a delayed reaction compared to natural behavior. To assess alignment between algorithm output and driver response, we compare the warning time ($t_w$) with the driver's natural reaction time without warning ($t_d$). Results show REACT's alerts align with human judgment and provide earlier warnings ($t_w < t_d$) in scenarios like car-following, rear-approaching, and intersection conflict.

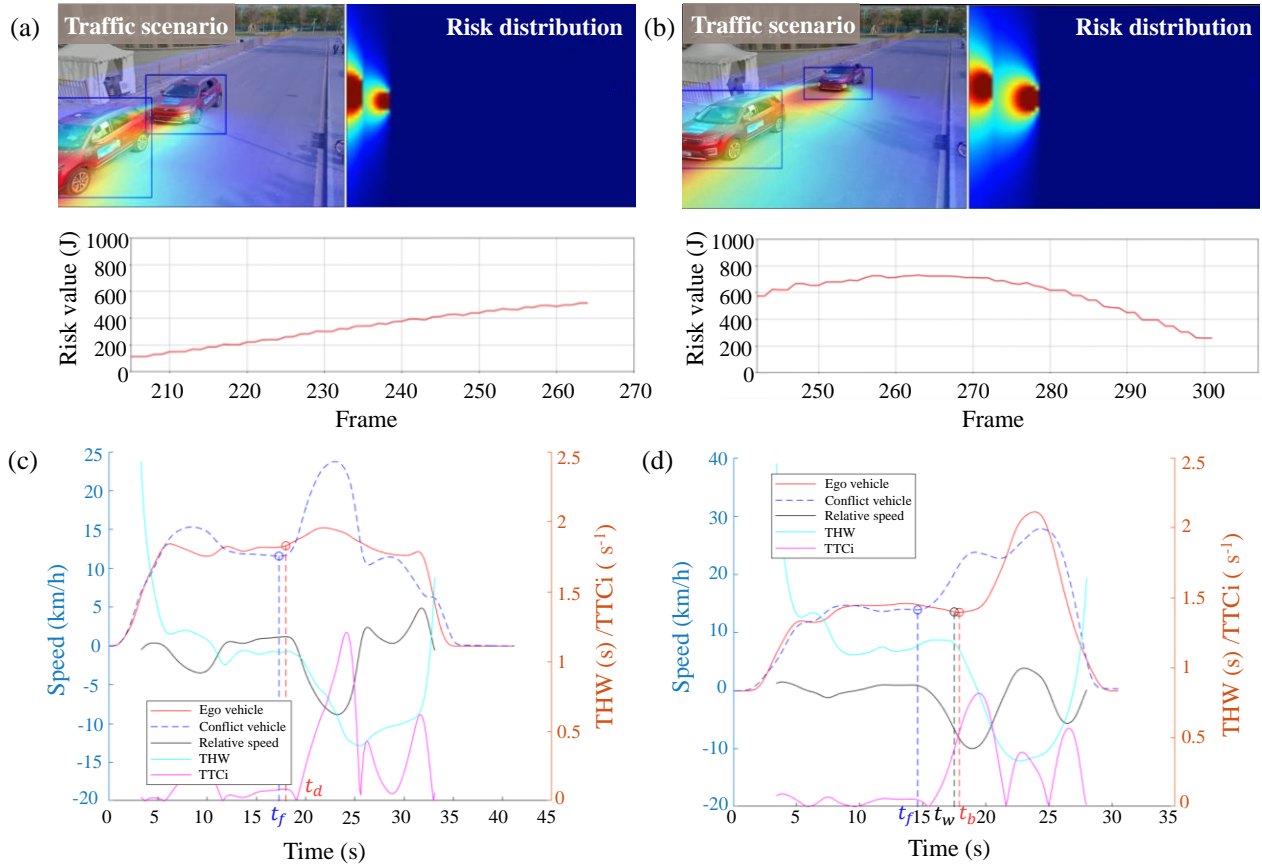

**Fig. 9.** Rear-approaching scenario: comparison with and without warning. (a) (c) Risk distribution and driver behavior without warning; (b) (d) Risk distribution and driver behavior with REACT-based warning.

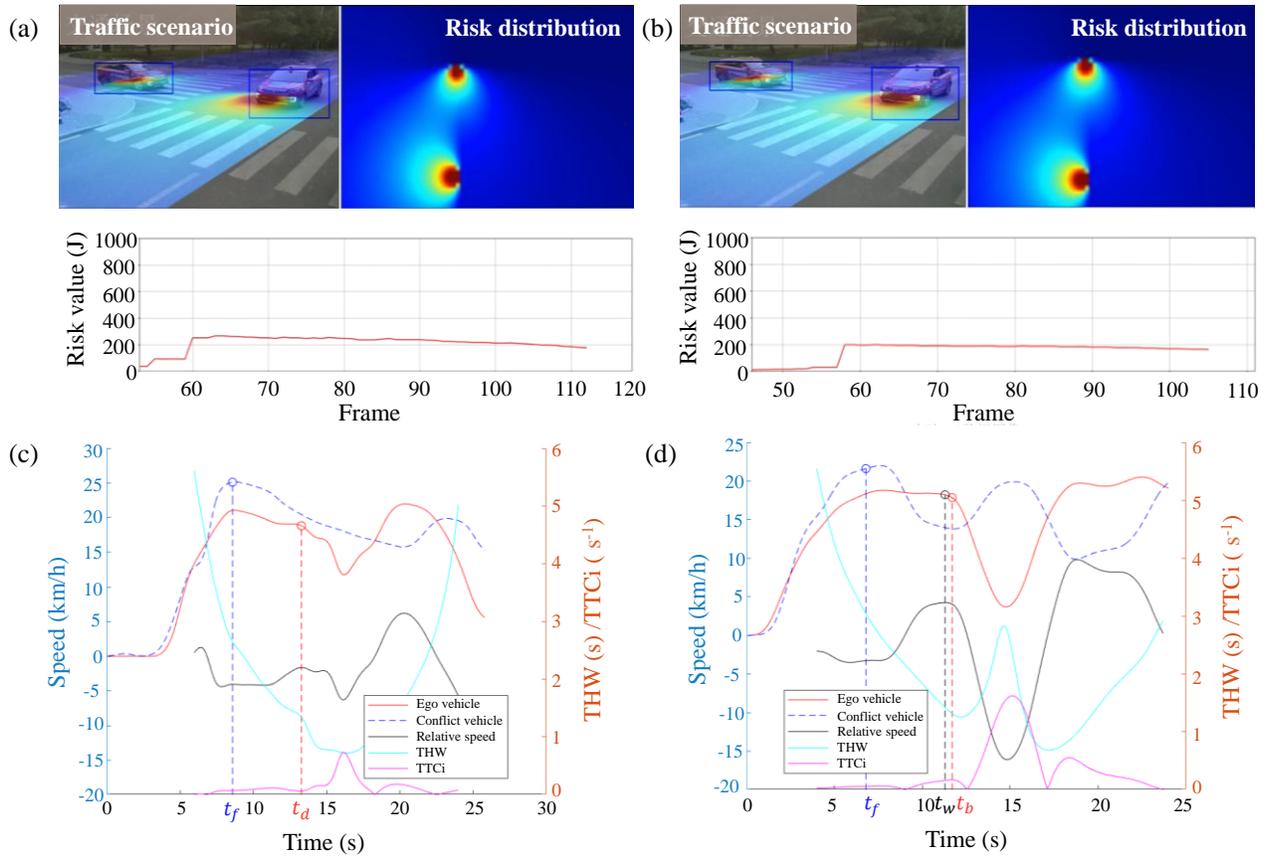

**Fig. 10.** Intersection conflict: comparison with and without warning. (a) (c) Risk distribution and driver behavior without warning; (b) (d) Risk distribution and driver behavior with REACT-based warning.

Fig. 11 further illustrates REACT's effectiveness in dynamic interactions. In (a), when a rear vehicle accelerates unexpectedly, the ego vehicle receives a forward acceleration warning and responds accordingly, achieving successful hazard avoidance. In (b), during a lateral deceleration-yielding conflict, the ego vehicle receives an alert and slows down to yield as the white vehicle accelerates into the lane.

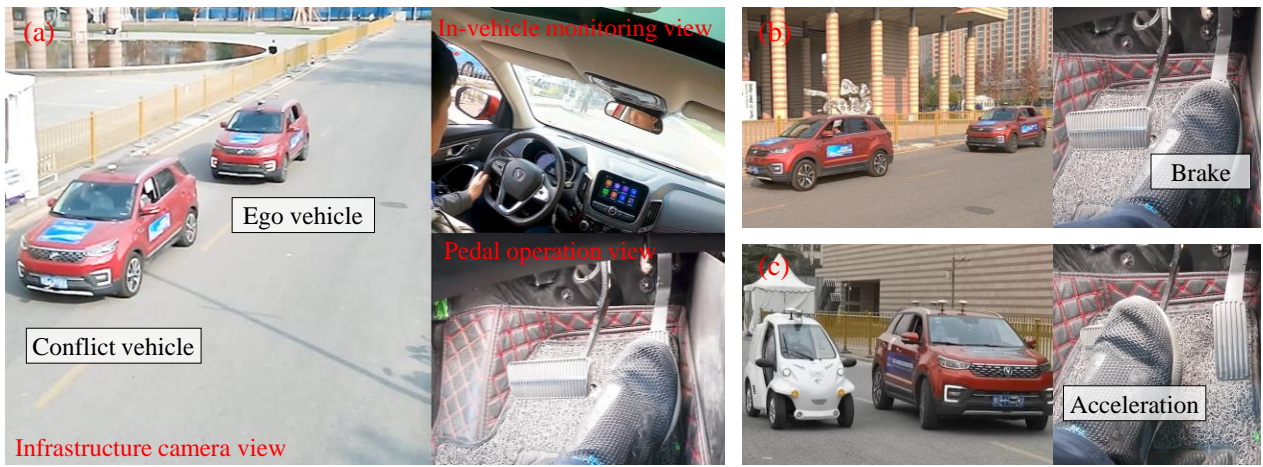

**Fig. 11.** Detected risk events during experiments. (a) Rear-approaching; (b) Lateral conflict with deceleration.

As shown in Table 4, driver responses under natural conditions and those triggered by REACT's risk warnings were analyzed across 16 scenario trials. CF, CI, RV, and IC denote car-following, cut-in, rear-approaching, and intersection conflict scenarios. Warning time is quantified as the time difference between the conflict vehicle's trigger time ($t_f$) and either the driver's braking time ($t_d$, yielding $\Delta t_{df} = t_d - t_f$) or the system's warning time ($t_w$, yielding $\Delta t_{wf} = t_w - t_f$). Specifically, in Table 4, the system achieved 0% false alarms, indicating no erroneous warnings during normal driving, and 0% misses, issuing timely alerts in all high-risk scenarios and enabling successful braking. In CF and IC scenarios, the time difference between the system warning ($t_w$) and the driver's natural braking moment ($t_d$) was less than 0.4 s, consistent with human anticipatory behavior and risk assessment. Additionally, the system's average response latency remained consistently around 35-50 ms across all scenarios, highlighting the superior real-time performance and execution efficiency of its lightweight architecture on embedded platforms.

Notably, in the RV scenario, false alarms and missed detections occurred in driver behavior. Specifically, drivers responded in non-hazardous situations with slight lead vehicle acceleration, while failing to act in truly risky cases. This explains why the average CI in multiple trials was recorded as 0.8 s, reflecting the driver's premature response, whereas REACT consistently issued accurate warnings with a CI of 2.7 s. The REACT system successfully alerted the driver to safely accelerate in every trial. Furthermore, the proposed method also accounts for vehicle interactions and the influence of intersection topology and geometry, enabling automatic adaptation to various intersection structures.

Table 4 Comparison between risk assessment algorithm and human risk perception behavior

| Risk assessment | Test | False alarm rate | Miss rate | Collision warning time | Collision prediction coverage |
|---|---|---|---|---|---|
| Human driver | 8 | 25% | 25% | CF = 2.7 s; CI = 0.6 s; RV = 0.8 s; IC = 4.8 s | Omnidirectional |
| REACT | 8 | 0% | 0% | CF = 2.3 s; CI = 1.7 s; RV = 2.7 s; IC = 4.4 s | Omnidirectional |

## 5. Conclusion

This paper introduces REACT, a unified, runtime-enabled collision avoidance framework that tightly integrates risk assessment with active control. Leveraging energy transfer principles and human-vehicle-road interaction modeling, REACT constructs a directionally adaptive, continuous risk field that enables precise threat quantification and interpretable avoidance behaviors. The grid-based directional risk attribution mechanism further enhances control efficiency by prioritizing collision-free paths under real-world constraints. In parallel, a lightweight hierarchical warning trigger strategy ensures physics-compliant, real-time responses. Comprehensive evaluations across four high-risk scenarios: car-following braking, vehicle cut-in, rear-approaching, and intersection conflict, demonstrate REACT's ability to achieve 100% safe avoidance with zero false or missed detections, a warning time consistent with human cognition (< 0.4 s), and a latency below 50 ms. These results confirm its superior foresight, generalization, and efficiency on embedded platforms, validating its readiness for safety-critical deployment. By bridging dynamic risk modeling with deployable system design, REACT offers a robust and scalable foundation for collision-free autonomous driving. Future work will further extend REACT's fully autonomous control and adaptability in unstructured environments, enhancing its applicability in real-world complex scenarios.

## CRediT authorship contribution statement

## Declaration of competing interest

The authors declare that they have no known competing financial interests or personal relationships that could have appeared to influence the work reported in this paper.

## Acknowledgements

We gratefully acknowledge the THICV research group at the School of Vehicle and Mobility, Tsinghua University, for providing the experimental platform, computational resources, and support from all participants involved in the on-road vehicle experiments.

## Data availability

Data will be made available on request.